\documentclass[10pt,twocolumn,letterpaper]{article}

\usepackage{cvpr}              

\usepackage[dvipsnames]{xcolor}

\definecolor{cvprblue}{rgb}{0.21,0.49,0.74}
\usepackage[pagebackref,breaklinks,colorlinks,citecolor=cvprblue]{hyperref}

\def\paperID{1954} 
\def\confName{CVPR}
\def\confYear{2024}

\title{FAAC: \underline{F}acial \underline{A}nimation Generation with \underline{A}nchor Frame and \underline{C}onditional Control for Superior Fidelity and Editability}

\usepackage{orcidlink} 

\author{Linze Li\orcidlink{https://orcid.org/0000-0001-9091-8699}  \and Sunqi Fan  \and Hengjun Pu \and Zhaodong Bing  \and Yao Tang  \and Tianzhu Ye \and Tong Yang \and Liangyu Chen \and Jiajun Liang$^\dagger$\orcidlink{https://orcid.org/0000-0001-5586-340X}  }
\begin{document}

\newcommand{\newteaser}{
	\centering
	\includegraphics[width=1.00\textwidth, trim=0em 0em 0em 0em, clip]{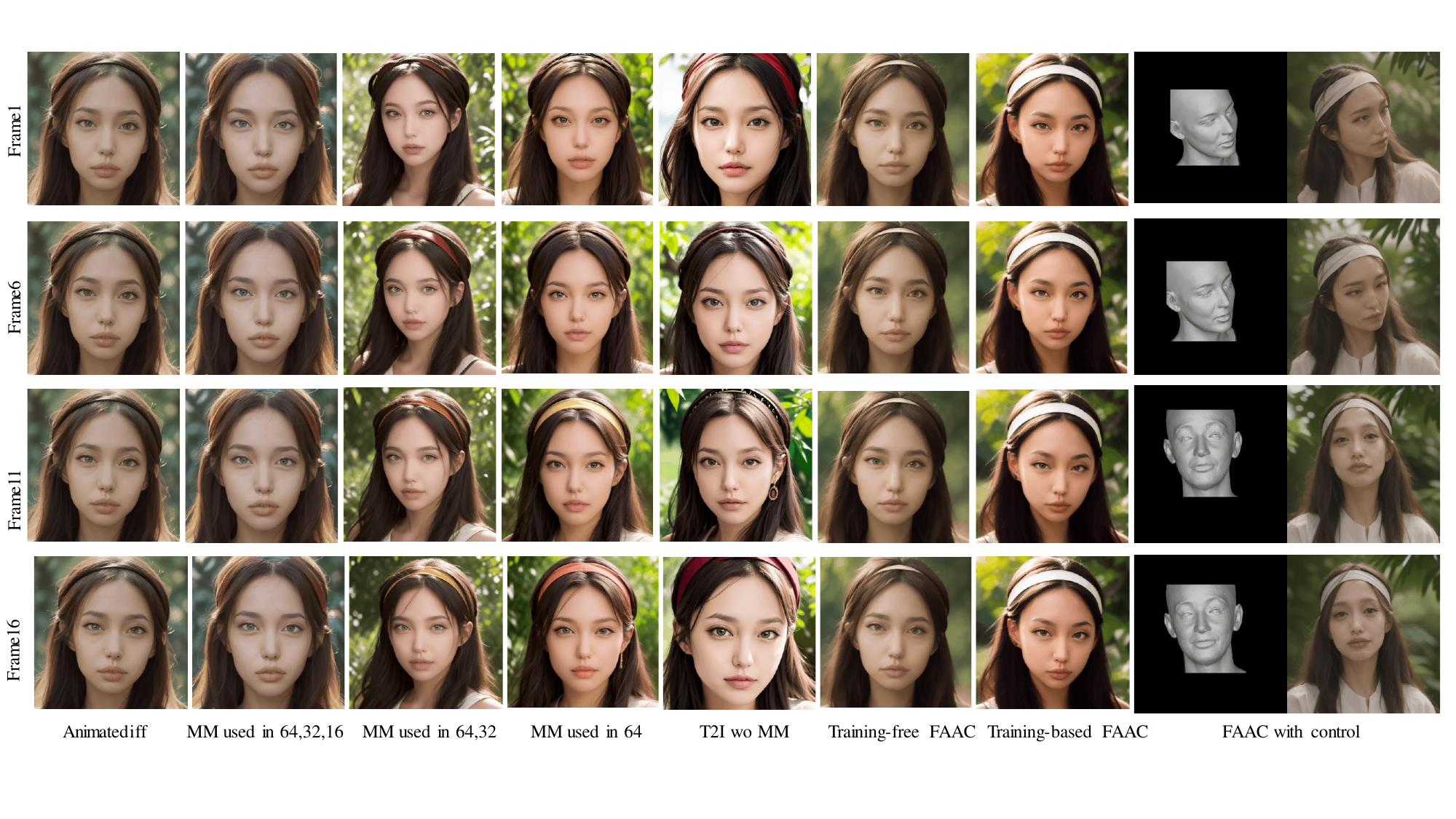}
    \vspace{-2em}
  \captionof{figure}{Our method provides three style high-fidelity and edited face animation: training-free FAAC, training-based FAAC and FAAC with conditional control. Compared with Animatediff\cite{guo2023animatediff} or its motion module in different resolutions, FAAC shows priority on the facial fidelity, text-to-image editability and video motion. MM stands for motion module, 64, 32, and 16 represent resolutions of 64x64, 32x32 and 16x16 respectively. T2I wo MM is the theoretical performance ceiling for facial fidelity and editing abality.
  \vspace{-1em}
  }
	\label{fig:main}
  \vspace{1.5em}
}

\twocolumn[{%
\vspace{-5em}
\maketitle%
\newteaser%
}]

\maketitle

\begin{abstract}
Over recent years, diffusion models have facilitated significant advancements in video generation. Yet, the creation of face-related videos still confronts issues such as low facial fidelity, lack of frame consistency, limited editability and uncontrollable human poses. To address these challenges, we introduce a facial animation generation method that enhances both face identity fidelity and editing capabilities while ensuring frame consistency. This approach incorporates the concept of an anchor frame to counteract the degradation of generative ability in original text-to-image models when incorporating a motion module. We propose two strategies towards this objective: training-free and training-based anchor frame methods. Our method's efficacy has been validated on multiple representative DreamBooth and LoRA models, delivering substantial improvements over the original outcomes in terms of facial fidelity, text-to-image editability, and video motion. Moreover, we introduce conditional control using a 3D parametric face model to capture accurate facial movements and expressions. This solution augments the creative possibilities for facial animation generation through the integration of multiple control signals. For additional samples, please visit our \href{https://paper-faac.github.io/}{project page}.

\end{abstract}    
\section{Introduction}
\label{sec:intro}
With the breakthrough development of deep learning and artificial intelligence, video generation technology has made significant progress in recent years. In particular, methods based on diffusion models have brought new opportunities to the field of video generation. Among these, facial animation generation, as an important branch of computer vision research, has attracted widespread attention due to its rich application scenarios, such as film production, virtual reality, and social media. High-quality facial animation generation needs to satisfy requirements of realism, expressive details, and ease of editing and control.

\begin{figure}[t]
    \centering
    {
    \includegraphics[width=0.45\textwidth, height=5cm]{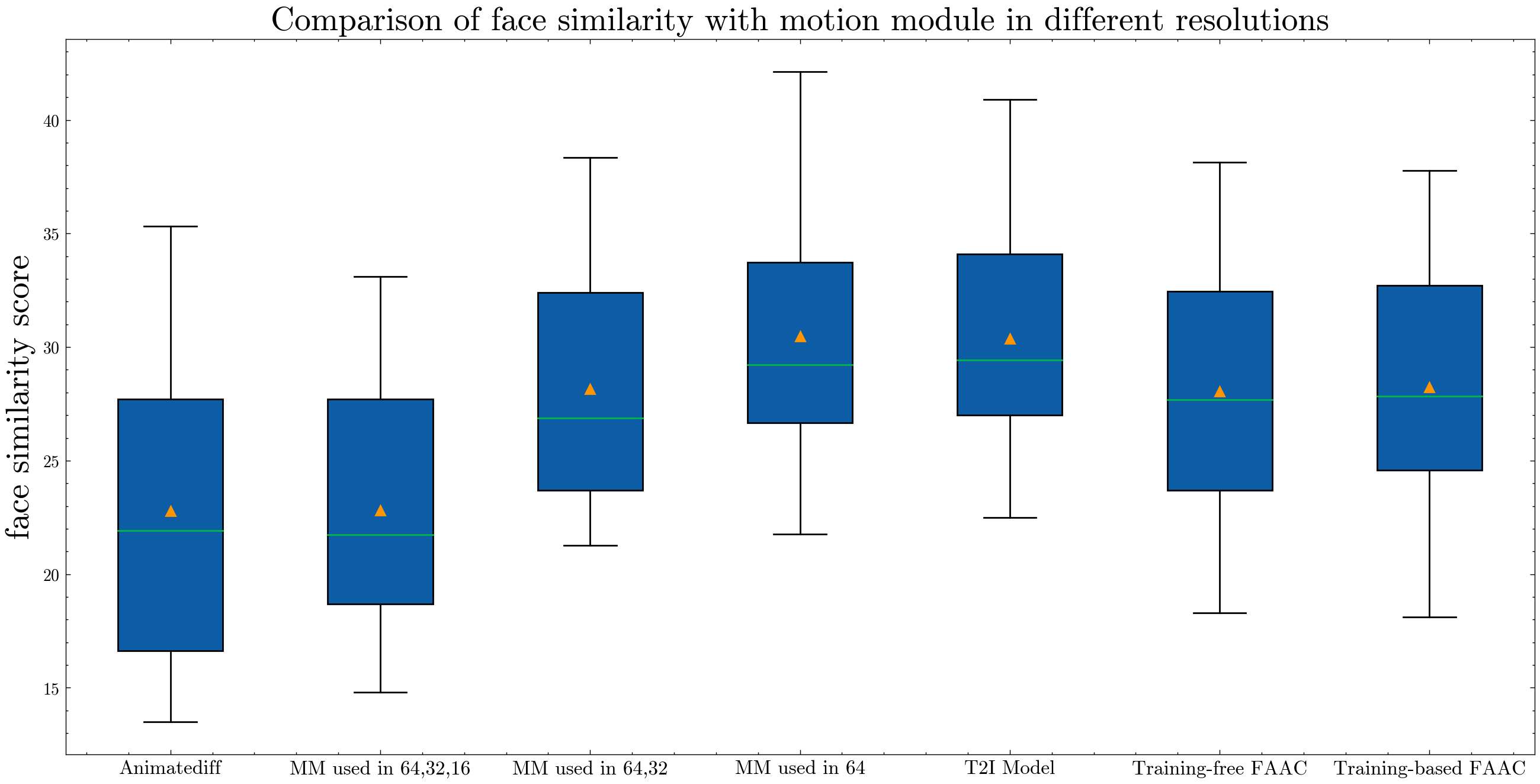}
    }\hfill
    {
    \includegraphics[width=0.45\textwidth, height=5cm]{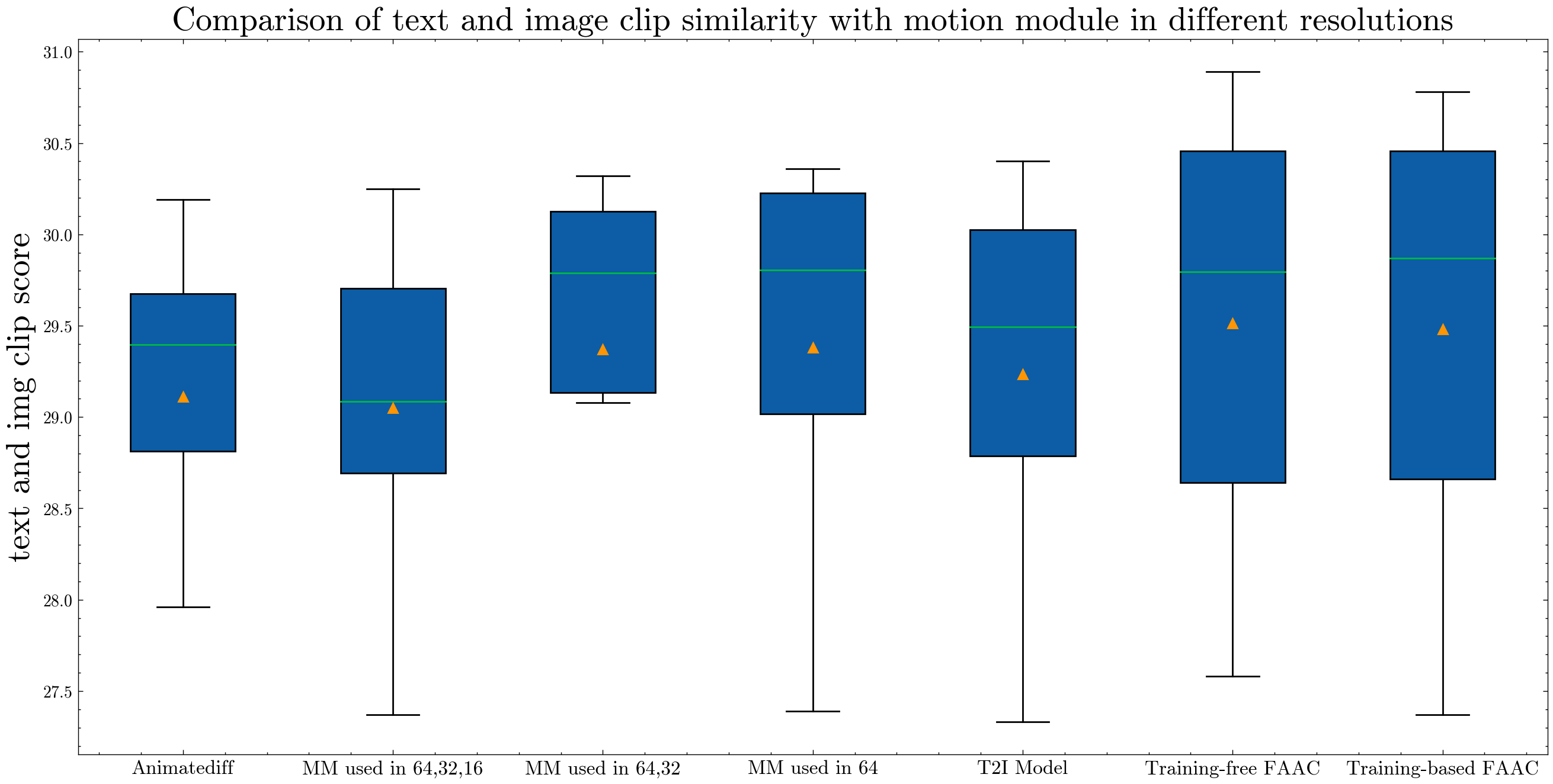}
    }
    \caption{Comparison of face similarity and text to image clip similarity with motion module in different resolutions and our methods
    }
    \label{fig:Comparison of face similarity and text to image clip}
\end{figure}

\begin{center}
\centering
\includegraphics[width=0.5\textwidth, trim=0em 0em 0em 0em, clip]
{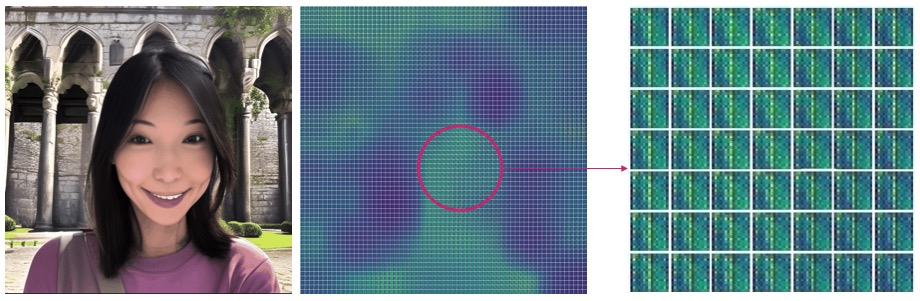}
\captionof{figure}{Visualization for Observation 2: Averaged attention maps from Temporal attention Module of each pixel. The model always attends to features of the same frame, causing the activation of a single column in the heat map. }
\label{observation2}
\end{center}
Despite this, a series of challenges still remain in high-quality facial animation generation, including realism, fidelity, and expressive detail. For example, although traditional face-swapping methods (such as Deepfakes\cite{Nguyen_2022}) can achieve a certain degree of realism, they are limited in terms of expression details and diversity, and they often struggle to maintain the fidelity of face identity with occlusions and pose changes. Also, their processing scope is limited to the facial area, making it impossible to effectively edit non-facial regions, thus reducing the creativity and editability of facial animation generation. Considering about creativity and editing of images or videos, the diffusion model, which has been particularly popular recently and has strong performance, comes to our mind. To our knowledge, the most popular and outstanding face or portrait video generation model based on the diffusion model is perhaps AnimateDiff\cite{guo2023animatediff}. The authors proposed a framework for extending any customized text-to-image model for animation generation, which can generate corresponding animation clips while maintaining the image style of the original customized model.

\begin{center}
\centering
\includegraphics[width=0.5\textwidth, trim=0em 0em 0em 0em, clip]
{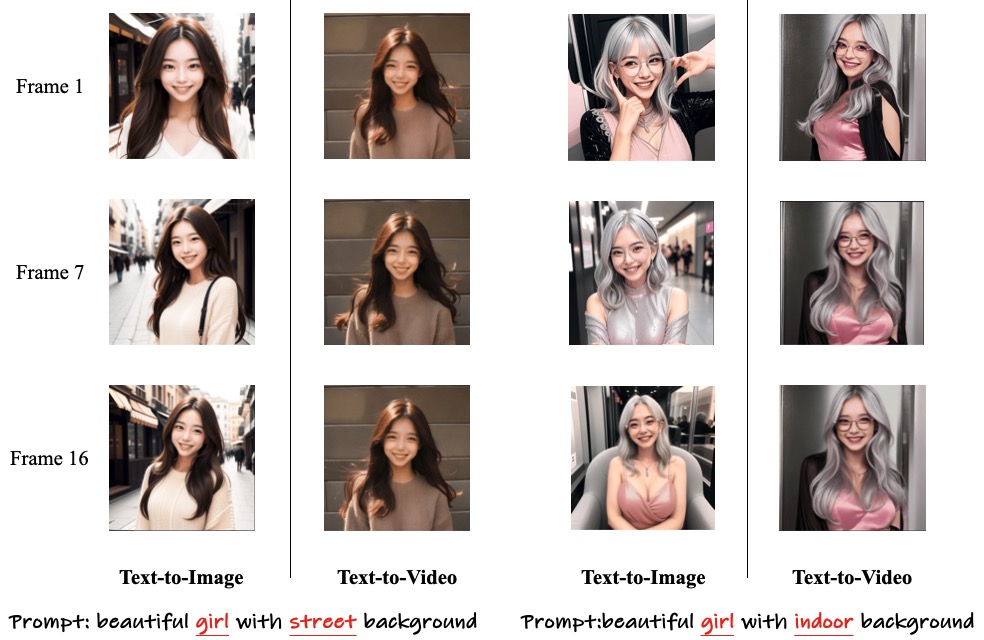}
\captionof{figure}{Visualization for Observation 3: Difference between T2I and T2V generated samples from same Gaussian noise start. Training for coherent frames will result in a simpler background and details.  }
\label{observation3}
\end{center}

We have three observations about the motion module of Animatediff\cite{guo2023animatediff} which inserted in the Stable Diffusion Model for video generation. \textbf{Observation 1: You can't have your cake and eat it too. Motion Module disrupts the generative ability of the original text-to-image model.} AnimateDiff\cite{guo2023animatediff} inserts a motion modeling module into each resolution of the text-to-image model. We found that the deep motion module disrupts the generative ability of the original text-to-image model. As shown in Figure \ref{fig:main}, as the layers deepen, the facial fidelity and editability of the model deteriorate, especially at the 16x16 and 8x8 stages. Figure \ref{fig:Comparison of face similarity and text to image clip} further quantifies the facial fidelity and editability of Animatediff, demonstrating a significant improvement when the deep motion model is discarded in the low resolution of latent. On the other hand, as motion module insert in the low resolution, the inter-frame continuity of video generation gradually strengthens. This raises the question: Is it possible to maximize the facial fidelity and editability to approach that of the text-to-image model without losing inter-frame continuity? \textbf{Observation 2: Keep an eye on the person ahead, makes the line move fast instead. Motion Module tends to align with the middle frame in the early stages of denoising.} As shown in Figure \ref{observation2}, we performed a visualization analysis of the temporal attention score maps and found that the model aligns with the middle frame already at timesteps 15. This reminds us that explicit modeling consistency to the anchor frame may help the model work better. \textbf{Observation 3: The shortest distance between two points is a straight line. Training for consistency frames makes generalized background simple and boring.} In order to meet the consistency constraints for frames, the model has a certain probability of favoring simpler backgrounds. As shown in Figure \ref{observation3}, we provided the prompt ``with street background'' and ``with indoor background'' independently, the original T2I model tends to generate complex street scenes and indoor scenes, while AnimateDiff\cite{guo2023animatediff} has an obviously higher probability of generating a simple wall and pure background.

In this work, we propose a novel facial animation generation scheme aimed at achieving realistic, smooth, high-fidelity, and richly detailed facial animations while enhancing the generation and editing capabilities of non-facial regions. We introduce the concept of an Anchor Frame to avoid the insertion of the motion model damaging the facial fidelity and editability of the text-to-image model based on observation 1 and observation 2. We provide both a training-free Anchor Frame inference method and a training-based Anchor Frame inference method. Both the training-free method and training-based method can improve both facial fidelity and editing capabilities by modeling the consistency with the anchor frame in the training process. Besides, in terms of facial fidelity and details, we introduce a conditional control using a 3D parametric face model to make the capture of facial movements and expressions more accurate.

To validate the effectiveness of the proposed method, we evaluate our AnimateDiff\cite{guo2023animatediff} on several representative DreamBooth and LoRA models about realistic photographs. Whether it's facial fidelity, text-to-image editability, or video motion, we have significantly improved compared to the original results. Additionally, we support combined conditional control generation and long-sequence video generation options, thus providing a broader creative space for facial animation generation.

The main contributions of this paper are as follows:
\begin{itemize}
\item We propose a novel facial animation generation method that aims to generate realistic, smooth, high-fidelity, and richly detailed facial animation while enhancing the generation and editing capabilities of non-facial regions. We introduce the training-free and training-based anchor frame method, which both counteract potential issues where the motion model might damage the facial fidelity and editability of the text-to-image model. 

\item We introduce a conditional control using a 3D parametric face model, making the capture of facial movements and expressions more accurate. Additionally, we support combined conditional control generation and long-sequence video generation options, thereby providing a broader creative space for facial animation generation.

\item We validate the effectiveness of our proposed method on multiple representative DreamBooth and LoRA models, and we have made significant improvements compared to the original results in terms of facial fidelity, text-to-image editability, or video motion. 
\end{itemize}

\section{Related work}
\label{sec:related work}


\subsection{Conditioned Video Generation with Diffusion Models}

Video Diffusion Models (VDM)\cite{ho2022video} first extended the DDPM\cite{ho2020denoising} models initially used for text-to-image generation, utilizing a factorized space-time U-Net to execute temporal attention based on the Text-to-Image (T2I) model. Then it becomes common to extend a T2I model with temporal structures for video generation. Animatediff\cite{guo2023animatediff} inserts motion modules into the U-Net to learn appropriate motion priors given textual descriptions. In addition, personalized generation can be achieved in Animatediff\cite{guo2023animatediff} by modifying the T2I model from base model (e.g., Stable Diffusion\cite{rombach2022high}) to personalized models, such as DreamBooth\cite{ruiz2023dreambooth} (utilizing a rare string) and Lora\cite{hu2021lora} (fine-tuning weights' residuals via low-rank decomposition). 

However, text descriptions often struggle to accurately represent complex motion. To enhance motion control within videos, some methods incorporate pose (e.g., Follow Your Pose\cite{ma2023follow}) or trajectories (e.g., DragNUWA\cite{yin2023dragnuwa}) to facilitate continuous video control. As a variety of conditions emerge, several works aim to develop frameworks that accommodate these diverse guidance. VideoComposer\cite{wang2023videocomposer} suggests categorizing conditions into three types (textual, spatial, and temporal) and employing condition fusion to leverage different control signals for collective guidance. Some works on image generation have also accomplished similar tasks. ControlNet\cite{zhang2023adding} employs a trainable copy of the initial neural network block with zero convolution layers, which receives extra conditions as input and adds the output to the original result. T2I-Adapter\cite{mou2023t2i} designs a simple adapter to extract multi-scale condition features. These methods can assist to synthesize videos since a T2I model can be extended to a Text-to-Video (T2V) model as previously mentioned.

\subsection{Face Generation with Diffusion Model}

Recently, diffusion models have been gradually applied to the highly practical domain of the human face. DiffSwap\cite{diffswap} firstly leverages the conditional inpaiting capability of diffusion models to perform face swapping task.To generate customized human portrait, MagiCapture\cite{hyung2023magicapture} follows DreamBooth\cite{ruiz2023dreambooth} to allocate special tokens for both source face and reference style, achieving disentanglement between face and background. Recent efforts have been increasingly focused on developing methodologies for generating animatable 3D-aware human facial models. Text2Control3D\cite{hwang2023text2control3d} leverages diffusion models with controlnet\cite{zhang2023adding} to generate multi-view face images, and then use these images to construct neural implicit field. Following DreamFusion\cite{poole2022dreamfusion}, FDNeRF\cite{zhang2023fdnerf} designs a diffusion loss to optimize the latent code representing a face, achieving a controllable prompt-driven face editing result. DreamFace\cite{zhang2023dreamface} generates animatable 3D face through three key steps: geometry generation, physically-based texture diffusion, and animation empowerment.

\begin{figure*}[t]
  \centering
\includegraphics[width=1.0\textwidth]{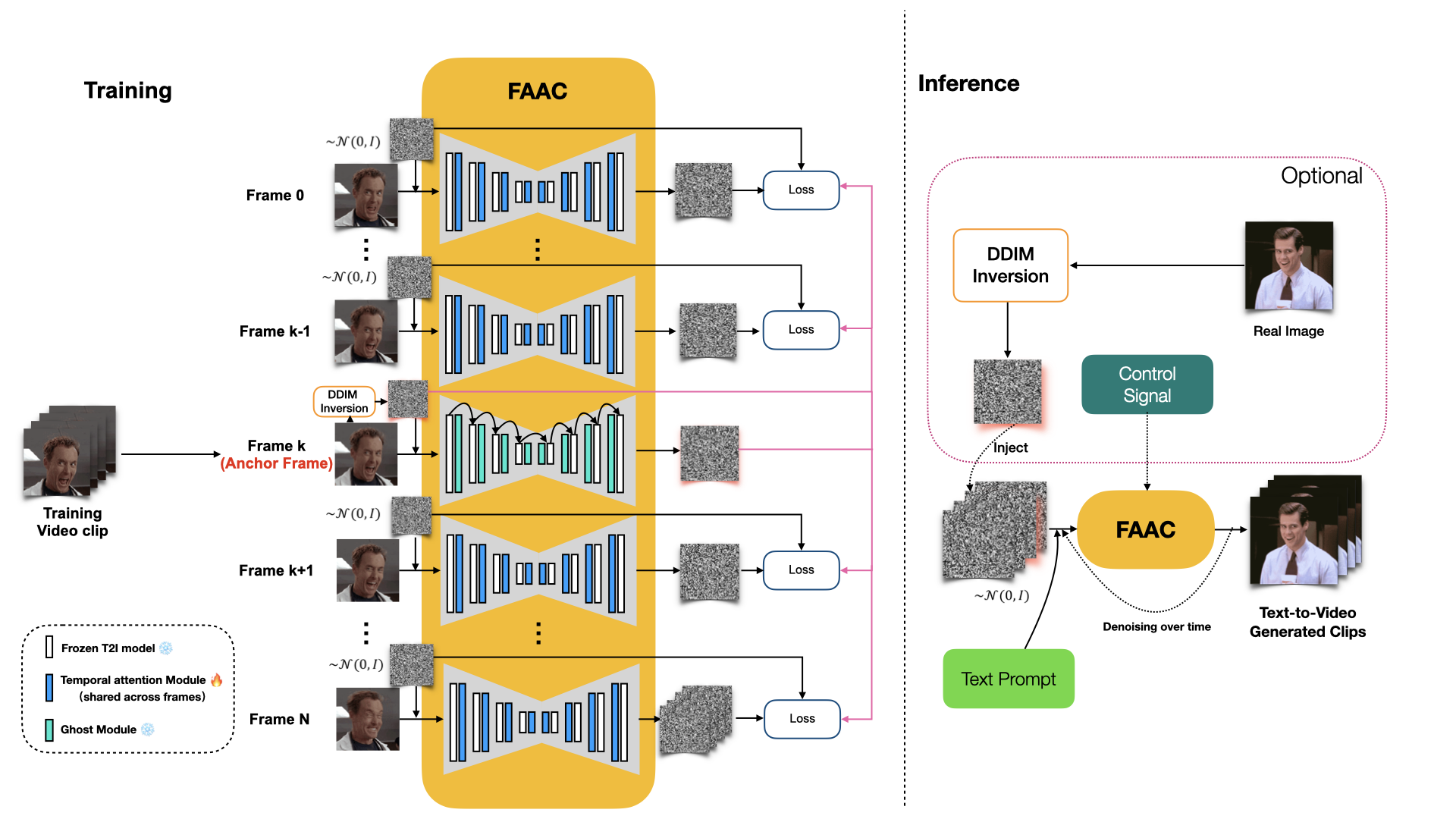}
  \caption{\textbf{Overview.} We propose a training method and an inference method for facial video generation. During training, we freeze the T2I model and train the Temporal attention Module with the help of our Anchor Frame mechanism. During inference, coherent video samples can be generated simply with our Anchor Frame Inference to get the same editability and fidelity of the original T2I model by the training-free or training-based approach.}
  \label{fig:teaser_method}
\end{figure*}

\section{Method}

As shown in Figure \ref{fig:teaser_method}, our method includes a frozen Text-to-Image diffusion model (T2I model)., a Temporal attention Module, and a Ghost Module. We first give a brief introduction to video generation from T2I models in \ref{subsec:T2I}. Training and inference schemes of our method FAAC are further discussed in \ref{subsec:anchorframetraining} and \ref{subsec:anchorframeinference}.
\subsection{Preliminary:Video Generation from Text-to-Image diffusion Models}
\label{subsec:T2I}
Diffusion models are generative models that generate samples that fit the real image distribution, recovering from totally random noise by a reversed denoising process. The forward process of diffusion models can be described as: $q_{t}(\emph{x}_{t} | {x}_{0})= \mathcal{N}(\emph{x}_{t}|\alpha(t)\emph{x}_{0},\beta(t)\emph{I})$, where $\emph{x}_0$ is a sample from original image distribution and $\alpha(t),\beta(t)$ are the noise scheduler weight. By estimating the noise and reversing the process, we can generate a novel sample by: 
\begin{equation}
\begin{aligned}
    p_{\theta}(\emph{x}_{0:T})&=p_(\emph{x}_T)\prod_{t=1}^{T} p_{\theta}(\emph{x}_{t-1}|\emph{x}_{t}) \\
    p_{\theta}(\emph{x}_{t-1}|\emph{x}_{t})&= \mathcal{N}(\emph{x}_{t-1};\mu_{\theta}(\emph{x}_{t},t),\sigma_{\theta}(\emph{x}_{t},t))
\end{aligned}
\end{equation}
where $\mu_{\theta},\sigma_{\theta}$ is estimated by neural networks $\theta$. The training objective of the diffusion model is the simple reconstruction loss: 
\begin{equation}
    L_{simple}(\theta)={||\epsilon_{\theta}(\emph{x}_{t},t)-\epsilon||}^2
\end{equation}
where $\epsilon$ is Gaussian random noise injected into a noisy sample $\emph{x}_t$ by noise scheduler.
Due to the success of diffusion models in capturing realistic images and video priors, 
there have been some excellent work of Text-to-Image diffusion models such as Latent Diffusion Models (LDM)\cite{rombach2022high}. LDM works on a lower resolution of latents so that the time efficiency and generation fidelity are well-balanced. There are two components in LDM: an autoencoder which encodes images and decodes the generated latents into images, and a normal diffusion model (modified U-Net architecture) which functions on noisy latents. 

The idea of turning image diffusion models into video diffusion models is very natural.  AnimateDiff \cite{guo2023animatediff} injects Temporal attention Module into the original LDM U-Net and trains the Temporal attention Module (cross-frame pixel-wise attention) on video datasets to generate text-to-video samples. It multiplexs the original T2I model's weight to process each single frame and uses the Temporal attention Module to get motion priors and objective consistency across frames. Our method followed AnimateDiff's architecture to use the Temporal attention Module. 

\subsection{Anchor Frame Motion Training}
\label{subsec:anchorframetraining}
To address the disadvantages brought by the Temporal attention Module in \textbf{Observation 1} and \textbf{Observation 3}, we propose Anchor Frame Motion Training to solve the problem. 

During training, we randomly choose a frame across all frames as Anchor Frame for each video batch. We annotate the anchor frame index as $k$.  the noisy latent of Anchor Frame is sent to the frozen T2I model and the Ghost Module. The Ghost Module does not modify any features of Anchor Frame so the output of predicted noise is the same as the output of the original frozen T2I model. The only function of Ghost Module is to send features of Anchor Frame to attend with features of other frames in the Temporal attention Module. The noisy latents of other frames are all sent to the frozen T2I model and Temporal attention Module. We only fine-tune the Temporal attention Module during training. The temporal attention can be expressed by: 
\begin{equation}
\begin{aligned}
    z_{i}&=Attention(Q_{i}K^{T})V ,  i\neq k \\
    Q_{i}&=W^{Q}z_{i},K=W^{K}z,V=W^{V}z \\
    z&=Cat(z_1,z_2,...,z_n)
\end{aligned}
\end{equation}
where $z_{i}$ is the feature of the i-th frame. The feature $z_{k}$ of Anchor Frame is not changed during the denoising process. Therefore it can fully utilize the editability and the fidelity of the T2I model and will not be affected by the performance degradation brought about by the motion module.

Due to the domain gap between training data for T2I models and Temporal attention Module (one is images, which we can hardly reach, and one is videos), cumulative errors may occur in the denoising process if the training targets of other frames during training are based on data from the video training set, but the anchor frame is not constrained. This could lead to a decrease in fidelity and abrupt alterations in the generated videos. 

To tackle this problem, we replace random noise with estimated noise from DDIM Inversion for the Anchor Frame. Therefore the predicted $x_0$ for Anchor Frame will be the nearly same as the original frame in the training video. The gap between the two domains can be effectively minimized by employing this approach, thereby facilitating the training of more coherent models.

\begin{center}
\centering
\includegraphics[width=0.3\textwidth, trim=0em 0em 0em 0em, clip]
{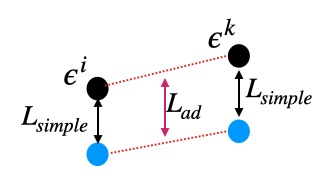}
\captionof{figure}{Graph of loss concept. Instead of learning to predict sampled random noise $\epsilon$, the loss will push the model to learn the difference between $\epsilon_{i}$ and $\epsilon_{i}$.}
\end{center}

\noindent \textbf{Training Objective.} Besides replacing the random noise with estimated noise from DDIM inversion, we redesigned the training loss (called Anchor Difference Loss) to diminish the gap instead of the simple reconstruction loss:
\begin{equation}
    L_{ad}=\sum\limits_{i=1, i\neq k}^{n}{||(\epsilon_{\theta}(\emph{x}_{t}^{i},t)-\epsilon_{\theta}(\emph{x}_{t}^{k},t))-(\epsilon^{i}-\epsilon^{k})||}^2/n
\end{equation}
where $\epsilon_{\theta}(\emph{x}_{t}^{i},t$ is the estimated noise from model of frame $i$ in at time step $t$ and $\epsilon^{i}$ is the sampled Gaussian noise of frame $i$.

The Anchor Difference Loss can decrease the latent gap between Anchor Frame and other frames, preserving the fidelity of faces. 
And the total loss of our training is $L=L_{simple}+\lambda L_{ad}$. We set $\lambda$ as 1.0 for our experiment.

\subsection{Anchor Frame Inference}
The inference pipeline is almost the same as the training procedure, as the anchor frame is generated only from the T2I model. The original $x_T$ of Anchor Frame can also be generated from any realistic image using DDIM Inversion, making our model suitable for animating any generated images or real images.

\noindent \textbf{Training-free.}We also discovered that the Temporal attention Module does not need to be fine-tuned to get quality results in our experiments. Therefore our inference method can also be training-free, which is a plug-and-play tool for facial video generation.

\noindent \textbf{Control signals.} Our method is also compatible with control signals to generate facial videos with certain facial landmarks, rendered face images from 3D Morphable Models, canny images, etc. See details in Section \ref{subsec:FACC With Control}. 
\label{subsec:anchorframeinference}


\section{Experiment}
In the experimental section, we firstly present the details of our training and evaluation in Section \ref{subsec:Experimental Detail}. Subsequently, in Section \ref{subsec:FAAC With Control}, we elucidate how we integrate FAAC with controllable generation through the utilization of 3D Morphable Model (3DMM)\cite{blanz2023morphable}. Following this, we discuss the qualitative and quantitative results of our method in Section \ref{subsec:Qualitative Results}, \ref{subsec:Quantitative Results}.

\label{sec:exp}
\subsection{Experimental Detail}
\label{subsec:Experimental Detail}
\noindent \textbf{Training.} We use Stable Diffusion v1.5 as our base text-to-image model to train the controllable modeling module of expression and pose and the motion module. We use 1000+ portrait videos as our dataset, the controllable modeling module was trained from scratch while the motion modeling module was fine-tuned with Animatediff V2 motion module. The video clips in the dataset are sampled at the stride of 1, then resized and center-cropped to the resolution of 512 × 512, the length of the video clips for training is set to 16 frames.

\noindent \textbf{Evaluations.} To confirm the efficacy and broad applicability of our approach, We use a diverse range of LoRA\cite{hu2021lora} collected from Civitai or trained by ourseleves, encompassing various genders, ages, and ethnicities. To enhance realism, following \cite{guo2023animatediff}, we also colleted a great number of realistic photography style DreamBooth\cite{ruiz2023dreambooth} from Civitai.
Our pipeline intergrates the base text-to-image model, motion module, stylized DreamBooth, and personalized LoRA, enabling the generation of highly realistic facial animations for the LoRA characters.

\subsection{FAAC With Control}
\label{subsec:FAAC With Control}
\noindent  
Based on our previous observations, Animatediff tends to prioritize ensuring inter-frame consistency in generated videos rather than inducing intricate motion. As a result, the generated video clips frequently display minimal movement in human actions, accompanied by negligible or no changes in facial expressions. The animation seems like a camera movement, and even common actions like blinking and smiling are infrequently observed.

In our pursuit of generating facial videos with more substantial motion, we leverage the power of T2I-adapter\cite{mou2023t2iadapter}. Similar to ControlNet\cite{zhang2023adding}, T2I-adapter is a commonly used method for controllable image generation given various conditions, such as pose, sketch, color, etc. Specifically, T2I-adapter utilizes a trainable Adapter structure to align internal knowledge in T2I models and external control signals.

The central focus of our investigation revolves around identifying a signal conducive to controlling the generation of facial animations. After careful deliberation and empirical exploration, our choice coalesces around the use of 2D rendering images obtained from a 3D Morphable Model (3DMM)\cite{blanz2023morphable, feng2021learning, Danek2022EMOCAED} as a conditional signal to control the generation of facial animations. Compared to facial landmarks, face parsing, and canny maps on human faces, using 2D rendering images of a 3D Morphable Model as condition offers several advanteages (demonstrated in Figure \ref{fig:multi-adapter}). 

\begin{center}
\centering
\includegraphics[width=0.45\textwidth, trim=0em 0em 0em 0em, clip]
{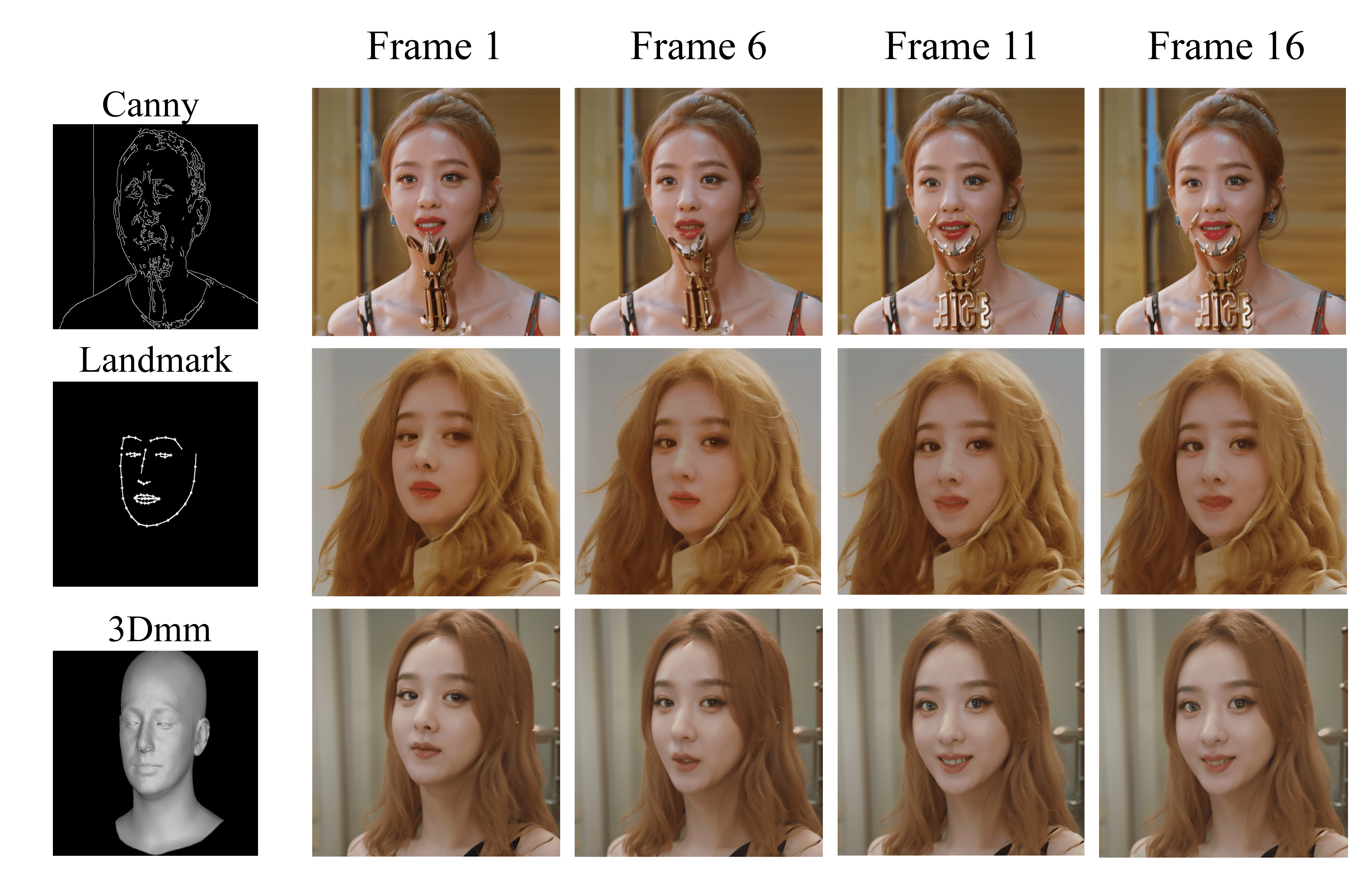}
\captionof{figure}{\textbf{The generation effects under different control signals.} It can be seen that the use of canny may result in disorderly lines that may not be correctly processed by Diffusion, thus failing to generate a reasonable facial video (the girl's chin in the first row). When generating using landmarks, it is sometimes challenging to accurately control the signal actions. Simultaneously, if the facial shape generated by the LoRA character differs significantly from that in the control signal, canny and landmark signal may lead to a decrease in fidelity. It can be observed that the use of 3DMM for generation has shown improvements in both fidelity and control accuracy of facial expressions.}
 \label{fig:multi-adapter}
\end{center}

\begin{figure*}[t]
    \centering
    \includegraphics[width=1.0\textwidth, trim=0em 0em 0em 0em, clip]{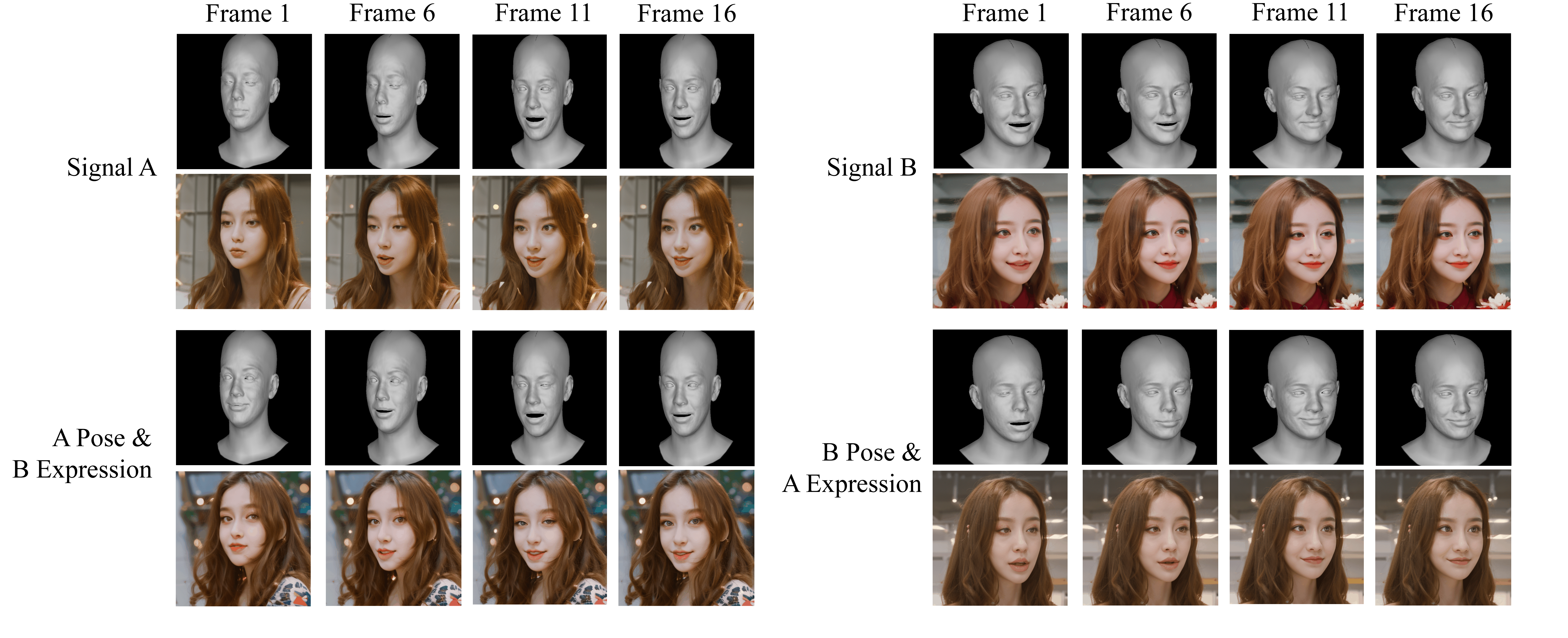}
    \captionof{figure}{\textbf{Example of pose and expression recombination.} It is worth noting that the pose feature also includes mouth features in 3DMM. Therefore, the party providing the pose feature determines the control signal for the mouth.}
    \label{fig:combination-fig}
\end{figure*}

Most importantly, we can recombine facial features with the assit of 3DMM. The 3DMM model enables the extraction of high-dimensional and fine-grained features such as pose, shape, expression, texture, and identity from a single image, condensed them into low-dimensional latent codes. By incorporating external features, such as the pose and expression of another individual, with the internal features of the facial shape, texture, and identity of the person we aim to generate, we enhance fidelity. Moreover, we can simultaneously capture facial features from multiple individuals. For instance, by incorporating the expression from individual A, the pose from individual B, and the facial shape and texture of the target-generated person. Figure \ref{fig:combination-fig} illustrates an example.


\subsection{Qualitative Results}
\label{subsec:Qualitative Results}

\begin{figure*}[t]
    \centering
    \includegraphics[width=1.0\textwidth, trim=0em 0em 0em 0em, clip]{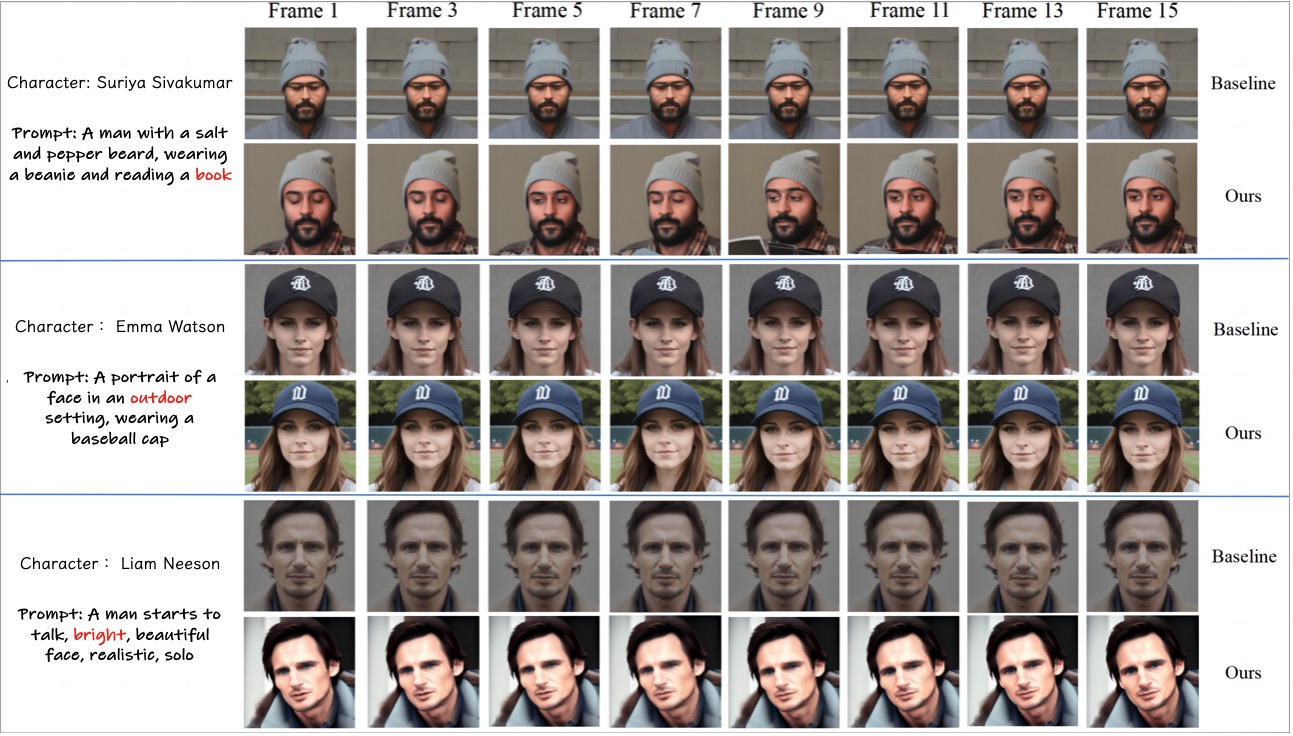}
    \captionof{figure}{\textbf{Qualitative Comparison.} Video sequence produced by our method (the second line in each unit) more closely align with the textual description and the LoRA character compared with the baseline (the first line in each unit).}
    \label{fig:qualitative-fig}
\end{figure*}

We conduct comprehensive evaluation under the same setting to fairly compare the qualitative results between our method and the AnimateDiff\cite{guo2023animatediff} baseline. We collected several typical and representative examples, as demonstrated in Figure \ref{fig:qualitative-fig}. It is worth noting the following three aspects:
\begin{itemize}
    \item Our approach is more in line with the prompt compared to the baseline. Descriptions incongruent with baseline samples are highlighted in red.
    \item Our method exhibits higher fidelity than the baseline. Under the same LoRA conditions, the facial animations generated by our approach more closely resemble the LoRA character.
    \item Compared to the baseline, our approach exhibits a greater range of motion. If control signals are incorporated, our samples can generate larger ranges of motion.
\end{itemize}

\subsection{Quantitative Results}
\label{subsec:Quantitative Results}
\noindent In order to comprehensively assess the fidelity, editability, and the overall quality of the generated facial animations, we use the following 3 kinds of metrics for qualitative evaluation.

\noindent \textbf{Face Similarity Score}: 
we leverage ArcFace\cite{Deng_2022} to assess the fidelity. ArcFace is a deep face recognition network which can extract representative face features through an Additive Angular Margin Loss. We employed ArcFace to extract facial embeddings from both the generated facial animations and authentic photographs of the LoRA character. The cosine similarity was computed to serve as the Face Similarity Score.

\noindent \textbf{CLIP Score}: we utilize the CLIP to assess the text-to-image editability. Specifically, we compute the CLIP\cite{radford2021learning} similarity between the prompt input of the diffusion model and the generated animation images to examine whether the prompt effectively controls and edits the generation of video content.

\noindent \textbf{Fréchet Video Distance}: following prior works\cite{skorokhodov2022styleganv, yan2021videogpt, ni2023conditional}, we utilize the Fréchet Video Distance (FVD) \cite{unterthiner2019accurate} to assess the overall quality of our generated video clips. FVD initially employs the pretrained I3D video classification network\cite{carreira2018quo} to extract feature representations from both real and synthesized videos. Subsequently, it computes the Fréchet distance between the distributions of features from real and synthesized videos. Following \cite{skorokhodov2022styleganv, xu2023pv3d}, we utilize CelebV-HQ\cite{zhu2022celebvhq} dataset as our real benchmark. And we use our approach and a baseline to synthesis a large number of videos under the same set of setting (prompts, DreamBooth, and LoRA) for fair comparison.


\begin{table}[!htp]\centering
\scriptsize
\caption{Quantitative Comparison}
\begin{tabular}{lrrr}\toprule
\textbf{Method} &  \textbf{Fidelity} ($\uparrow$) & \textbf{Editability} ($\uparrow$) & \textbf{FVD} ($\downarrow$)\\\midrule
AnimateDiff & 22.79& 29.11&595.45\\
training-free FAAC & 28.04&29.48&\textbf{504.15}\\
training-based FAAC & \textbf{28.24}&\textbf{29.51}&523.27\\
\bottomrule
\end{tabular}
\end{table}

Under the above three metrics, we compare our methods with AnimateDiff\cite{guo2023animatediff}. We observed a significant improvement in fidelity, editability, and overall video quality for both training-free and training-based FAAC compared to AnimateDiff. While the training process may hurt the FVD metric, it concurrently enhances fidelity and editability.

\section{Limitations and Future Works}

Despite the promising results achieved by our method, we also recognize several limitations and potential avenues for future exploration.

In our anchor frame approach, we observed that the image quality of non-anchor frames is lesser compared to anchor frame. The denoising process resembles a chasing problem: despite striving to match the anchor frame, it never reaches perfection. Furthermore, in Observation 2, we mentioned that inserting a motion model at a deep level into the T2I model results in loss of the original facial fidelity and editability provided by the T2I model. However, our current approach has not yet capitalized on this insight, marking an area for future investigation.


Another significant aspect of our research accentuates the importance of conditional control using a 3D parametric face model. This feature enables us to capture facial movements and expressions more accurately, contributing significantly to our method's performance. However, we found that introducing control signals, such as 3DMM, harms the original fidelity of the T2I model—a challenge worth continuing research. Moreover, how to generate control signals from text is a research topic in anticipation of accomplishing comprehensive fine-motion video generation merely using text.

Additionally, our current facial fidelity is entirely sourced from the LoRA model. Exploring other methods to generate videos of specific ID faces is also a direction for our future work.

\section{Conclusion}

In this work, we proposed a novel facial animation generation approach, using the potential of diffusion models enriched with our newly introduced anchor frames and conditional control. Our technique effectively handles previously observed challenges in fidelity and editability when incorporating motion dynamics into Text-to-Image (T2I) models.

The integration of a 3D parametric face model added to our methodology by providing a more accurate capture of facial movements and expressions, contributing to the dynamic realism and the 3D-consistency of the generated animations. Furthermore, our method's capacity for combined conditional control generation opened up new possibilities for creative applications.


Our experimental results validated our method's effectiveness on multiple representative DreamBooth and LoRA models, showing significant improvements in facial fidelity, text-to-image editability, and video motion compared to existing solutions.

{
    \small
    \bibliographystyle{ieeenat_fullname}
    \bibliography{main}
}


\end{document}